\documentclass{article}
\usepackage{spconf,amsmath,epsfig}
\usepackage{amsfonts}
\usepackage{amssymb}
\usepackage[font=small,labelfont=bf]{caption}
\usepackage{float} 
\usepackage{caption} 
\usepackage{booktabs}
\usepackage{multirow}
\usepackage{xcolor}
\title{MIMRS: A S\lowercase{urvey on} M\lowercase{asked} I\lowercase{mage} M\lowercase{odeling} \lowercase{in} R\lowercase{emote} S\lowercase{ensing}}

\name{Shabnam Choudhury$^{1}$, Akhil Vasim$^{1}$, Michael Schmitt$^{2}$, Biplab Banerjee$^{1}$ \thanks{Corresponding author: choudhury.shabnam6@gmail.com}}
\address{$^{1}$Centre of Studies in Resources Engineering, Indian Institute of Technology, Bombay, India \\
$^{2}$Department of Aerospace Engineering, Bundeswehr University Munich, Germany}

\begin{document}
\maketitle
\begin{abstract}
Masked Image Modeling (MIM) is a self-supervised learning technique that involves masking portions of an image, such as pixels, patches, or latent representations, and training models to predict the missing information using the visible context. This approach has emerged as a cornerstone in self-supervised learning, unlocking new possibilities in visual understanding by leveraging unannotated data for pre-training. In remote sensing, MIM addresses challenges such as incomplete data caused by cloud cover, occlusions, and sensor limitations, enabling applications like cloud removal, multi-modal data fusion, and super-resolution. By synthesizing and critically analyzing recent advancements, this survey (MIMRS) is a pioneering effort to chart the landscape of mask image modeling in remote sensing. We highlight state-of-the-art methodologies, applications, and future research directions, providing a foundational review to guide innovation in this rapidly evolving field.
\end{abstract}
\begin{keywords}
Masked image modeling, self-supervised learning, masked autoencoder,
\end{keywords}
\vspace{-5pt}
\section{Introduction}
\vspace{-5pt}
The rapid proliferation of Earth observation satellites has enabled the collection of vast amounts of unlabeled remote sensing imagery. Despite this abundance, many remote sensing models continue to initialize with ImageNet \cite{deng2009imagenet} pre-trained weights. This method is effective for natural images but is significantly constrained when used for remote sensing tasks. The domain disparity between natural and remote sensing images marked by variations in spatial resolution, spectral bands, and scene composition limits the applicability of ImageNet-trained features to remote sensing data. This inconsistency leads to models with inadequate generalization ability. Additionally, the need for large-scale labeled data further hinders the scalability of models trained using ImageNet pre-training. Remote sensing data sets often require domain-specific annotations, such as land cover classifications or temporal changes, which are labor-intensive and costly. This dependence on continuous manual annotation undermines the efficiency of practical applications and creates bottlenecks in developing generalized RS models. They exhibit diverse spatial resolutions and orientations due to the variability in RS sensors and the aerial viewpoint, leading to significant angular and scale variations for identical objects. Additionally, RS images often contain densely packed, small objects distributed over large areas, which complicates object detection and interpretation compared to natural images that typically include fewer objects.  \par

 Consequently, there is a growing need to explore self-supervised representation learning tailored specifically to remote sensing data, which can better bridge this domain gap and unlock the full potential of the available data. In recent years, self-supervised learning (SSL) has emerged as a dominant paradigm for pre-training models in computer vision \cite{bao2021beit, chen2020simple, he2020momentum, he2022masked}. This paradigm has gained significant traction in remote sensing (RS) and Earth observation \cite{wang2022self} due to two primary factors. First, many high-impact RS applications, including crop yield estimation, urban planning, and disaster management, are constrained by the scarcity of labeled data \cite{christie2018functional, schmitt2019sen12ms, sumbul2019bigearthnet}, despite the abundance of unlabeled satellite imagery. SSL offers a practical solution by leveraging this enormous pool of unlabeled data to learn robust representations without requiring extensive manual annotations. \par

\begin{figure}
    \centering
    \includegraphics[height = 3.2cm,width = 8.5cm] {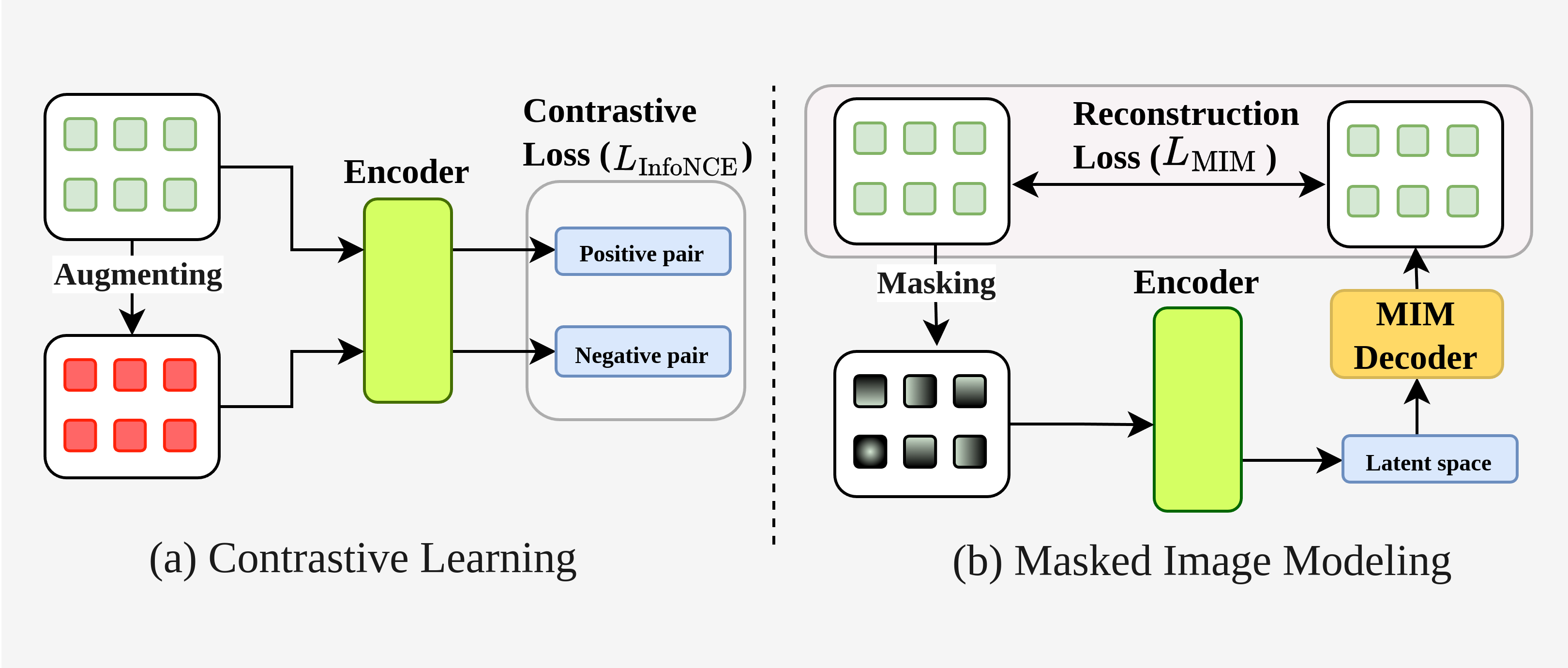}
    \caption{Representation of two prominent self-supervised methods.}
    \label{Figure1}
\end{figure}

\begin{figure*}
    \centering
    \includegraphics[height = 7.2cm,width = 17cm]{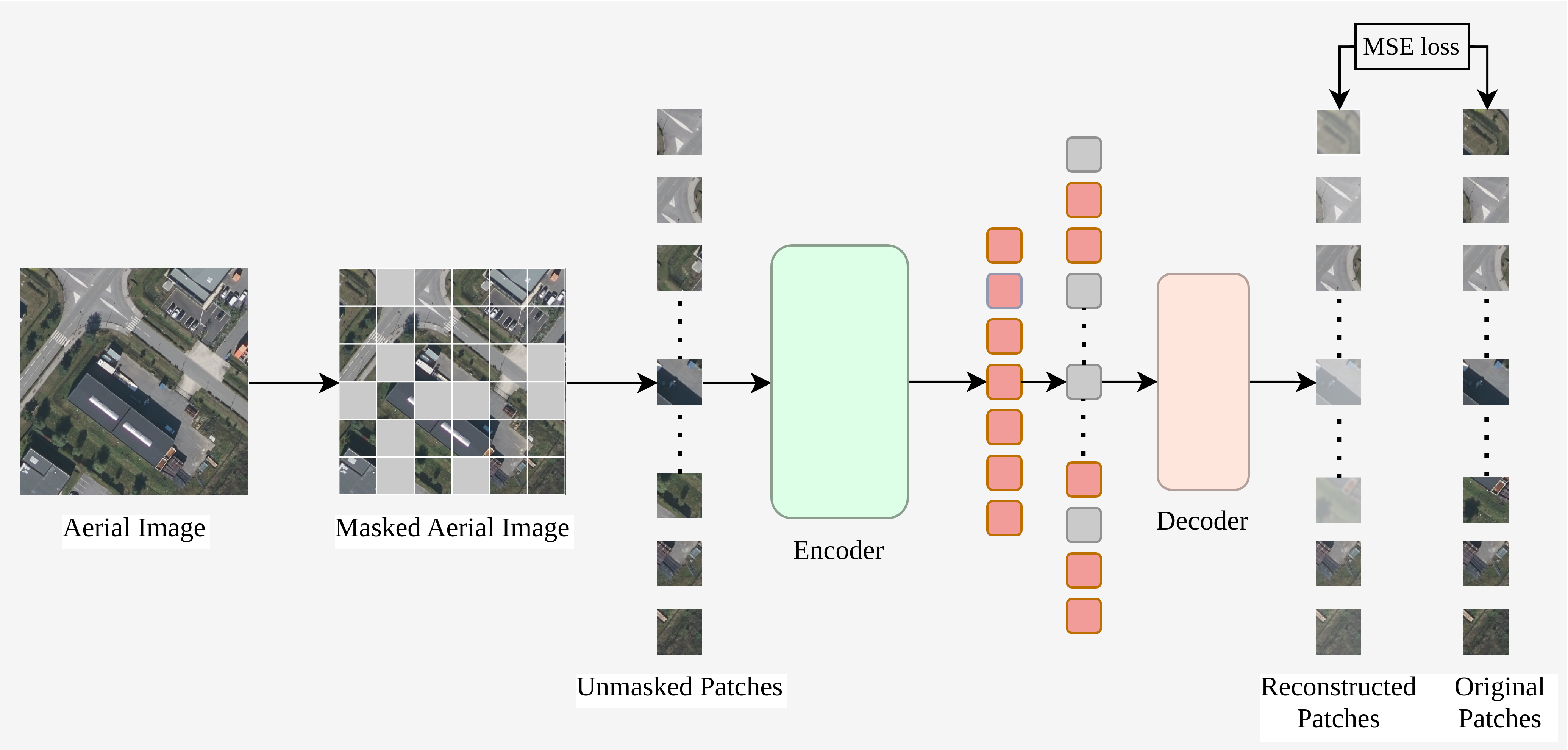}
\caption{Illustration of a fundamental framework proposed by MAE \cite{he2022masked} for MIM pretraining, in which the visible patches are encoded, and the encoded features are decoded in conjunction with masked patches to reconstruct the pixel}
    \label{Figure2}
\end{figure*}
 In recent years, contrastive learning  \cite{chen2020simple, he2020momentum} has prevailed among SSL approaches in the domain of remote sensing. The core idea of contrastive learning is to bring similar samples closer in the feature space while pushing dissimilar samples apart. Without explicit labels, positive pairs are typically created through data augmentation of the same image, while other images serve as negative pairs. This method has proven particularly efficacious in remote sensing, where data augmentation methods \cite{ayush2021geography, manas2021seasonal} like rotation, cropping, and spectral alterations may produce varied perspectives of the same sample. To enhance contrastive learning, domain-specific information such as geographic metadata \cite{oord2018representation}, temporal sequences \cite{akiva2022self}, and additional inputs like acoustic or environmental data \cite{manas2021seasonal} have been incorporated. SkySense \cite{guo2024skysense} pioneered a contrastive learning methodology that integrates many modalities and geographical scales, employing geo-contextual prototypes to provide efficient cross-modal integration of multispectral (MSI), RGB, and SAR data. Despite its success, contrastive learning methods often assume that different images inherently belong to distinct classes. Nevertheless, with RS datasets, this assumption is often inaccurate since multiple images may have instances of the same class, imposing possible constraints on the method's efficacy in certain contexts.
 
 To address these limitations, Masked Image Modeling (MIM) \cite{bao2021beit,he2022masked, xie2022simmim} has emerged as a powerful alternative within SSL paradigms for learning visual representations. Unlike contrastive learning, MIM operates by masking portions of an input image, such as individual pixels, image patches, or latent features, and training models to reconstruct the missing parts based on the visible context. This shift enables MIM to utilize the inherent structure and redundancy in image data, resulting in more robust and generalizable feature representations. By directly learning from the image itself without assumptions about inter-image relationships, MIM circumvents the drawbacks of contrastive learning and adapts seamlessly to the unique challenges of remote sensing data. The simplicity of MIM architectures, combined with their remarkable performance on downstream tasks, has garnered significant attention in the computer vision community \cite{he2022masked, li2021mst, xie2022simmim}. Recent advancements have extended the application of MIM to remote sensing imagery. Studies have shown that MIM can serve as an effective pre-training strategy for vision transformers \cite{dosovitskiy2020image, liu2021swin}, which benefit from MIM’s ability to capture spatial and contextual patterns in remote sensing data \cite{cong2022satmae, reed2023scale, wang2022advancing}. These findings underscore the potential of MIM to enhance representation learning for remote sensing tasks, paving the way for improved performance in applications such as classification \cite{wang2023remote} segmentation \cite{qiao2023semask}, and change detection \cite{li2023segmind}.\par

 In this paper, we provide a cohesive perspective of Masked Image Modeling (MIM) that categorizes its diverse applications across remote sensing tasks and modalities and highlights its transformative potential in addressing domain-specific challenges.

\begin{table*}
\centering
\begin{center}
{%
\begin{tabular}{lllcccccc}
\toprule
\multirow{3}{*}{Model} & \multirow{3}{*}{Backbone} & \multirow{3}{*}{Params (M)}
& \multicolumn{2}{c}{\textit{Scene Classification}} & \multicolumn{2}{c}{\textit{Object Detection}}& \multicolumn{2}{c}{\textit{Semantic Segmentation}}\\
& & & {AID} & {RESISC-45} &{DIOR} & {DIOR-R} & {LoveDA} & {SpaceNetV1} \\
& & & {TR=50\%} & {TR=20\%} & {mAP$_{50}$} & {mAP$_{50}$} & {mIoU} & {mF1} \\
\midrule
\textbf{\textit{Contrastive - based}}\\
SeCo \cite{manas2021seasonal} & ResNet-50 & 26 & 95.99 & 92.91 & -- & -- & 43.63 & 77.09 \\
GASSL \cite{ayush2021geography} & ResNet-50 & 26 & 95.92 & 93.06 & 67.40 & 65.65 & 48.76 & 78.51 \\
CaCo \cite{mall2023change} & ResNet-50 & 26 & 95.05 & 91.94 & 66.91 & 64.10 & 48.89 & 77.94 \\
\textbf{\textit{MAE - based}}\\
SatMAE \cite{cong2022satmae} & ViT-L & 307 & 96.94 & 94.10 & 70.89 & 65.66 & -- & 78.07 \\
ScaleMAE \cite{reed2023scale} & ViT-L & 307 & 97.58 & 95.04 & 73.81 & 68.17 & -- & -- \\
SSL4EO \cite{wang2023ssl4eo}& ViT-S & 22 & 94.82 & 91.21 & 67.91 & 61.23 & -- & -- \\
RingMo \cite{sun2022ringmo}  & Swin-B & 88 & 95.06 & 95.06 & 75.90 & 67.59 & -- & -- \\
RVSA \cite{wang2022advancing}  & ViT-B+RVSA & -- & {98.50} & 95.69 & 75.80 & 70.51 & {54.00} & {54.00} \\
SelectiveMAE \cite{wang2024scaling}
& ViT-L & 307 & 98.48 & 95.77 & {77.80} & 77.80 & {54.31} & {79.46} \\
\bottomrule
\end{tabular}}
\caption{Performance of contrastive-based and MAE-based models on various downstream tasks}
\label{table1}
\end{center}

\vspace{-0.5cm}
\end{table*}
\vspace{-5pt}

 \section{Generic Framework}
 \vspace{-5pt}
 \subsection{Overview}
 \vspace{-0.2cm}
 \noindent \textbf{A. Masked Image Modeling (MIM)}: Drawing inspiration from the success of Masked Language Modeling (MLM) in natural language processing \cite{devlin2018bert}, MIM has been developed as a powerful framework for visual pre-training \cite{he2022masked, xie2022simmim}. It aims to learn robust image representations by reconstructing masked tokens, employing a variety of regression targets \cite{bao2021beit, he2022masked}, innovative masking strategies \cite{li2022semmae}, and diverse reconstruction methods \cite{li2021mst}. As a case study, BEiT \cite{bao2021beit} masks 60\% of an image and employs tokens extracted through discrete variational autoencoders (dVAE) \cite{rolfe2016discrete} to forecast the masked regions. SimMIM \cite{xie2022simmim} simplifies the procedure by directly forecasting pixel values of masked patches, incorporating both visible and masked tokens. Although MIM is highly effective, it frequently encounters obstacles, such as extended pre-training periods and a high level of computational complexity. 
 MAE enhances computational efficiency by exclusively processing the visible regions within the encoder \cite{he2022masked}. CrossMAE \cite{fu2024rethinking} adopts an alternative methodology by utilizing cross-attention mechanisms between masked and visible tokens, hence improving efficiency without compromising performance. Although these developments have greatly expanded the scope of self-supervised representation learning in natural image domains, they frequently neglect the unique constraints posed by remote sensing (RS) images. This necessitates the customized modifications of MIM to maximize its efficacy in remote sensing applications.

 \vspace{0.7mm}

  \noindent \textbf{B. Remote Sensing Masked Image Modeling (MIM-RS)}: Visual representations in natural images have been effectively learned through the use of MIM. Nonetheless, adopting MIM for remote sensing imaging necessitates tackling distinct issues inherent to this field. Conventional self-supervised techniques reliant on contrastive learning \cite{ayush2021geography, mall2023change, wanyan2023dino} frequently encounter inefficiencies in the formulation of pretext problems and the acquisition of labeled data, resulting in an increasing emphasis on generative self-supervised methods, notably MIM. One approach \cite{wang2022advancing} involved pre-training the MAE \cite{he2022masked} framework on the MillionAID dataset and enhancing its performance on downstream tasks by substituting the traditional global attention mechanism in transformers with rotational and variable-size window attention. Another method, CMID \cite{muhtar2023cmid} improves the task by integrating contrastive learning components to ensure consistency in the learnt representations. RingMo \cite{sun2022ringmo} employs a novel patch-based imperfect masking algorithm tailored for the reconstruction of satellite and aerial photos, utilising a dataset of over two million images. Certain methodologies, like GFM \cite{mendieta2023towards}, employ GeoPile, a comprehensive dataset aggregated from many sources, for pre-training. These methods enhance in-domain feature representations by ongoing learning directed by models pre-trained on extensive datasets, like ImageNet-22k. SatMAE \cite{cong2022satmae} integrates temporal and multi-spectral data into position embeddings, allowing it to proficiently capture spatio-temporal correlations, as demonstrated in applications with fMoW \cite{christie2018functional} datasets. ScaleMAE \cite{reed2023scale} utilizes ground sample distance to rebuild images at various resolutions, leveraging the intrinsic multi-scale characteristics of remote sensing data.\par

  In addition to geographical dimensions, masking algorithms have been applied to the spectral domain, with application in multispectral (MSI) \cite{hong2024spectralgpt}, hyperspectral (HSI) \cite{wang2024hypersigma}, and multimodal data processing. Furthermore, investigations have examined multimodal MIM frameworks for paired remote sensing data corresponding to geographic locations. For example, msGFM \cite{han2024bridging} implemented a multimodal design that includes a common encoder and separate patch embedding and decoder modules, allowing it to handle various data formats, including RGB, MSI, SAR, and DSM. Likewise, MMEarth \cite{nedungadi2024mmearth} established a multimodal MAE framework intended to manage 12 pixel-level and image-level modalities, demonstrating the viability of MIM in sophisticated multimodal remote sensing applications.

\vspace{-5pt}
 \subsection{SSL Pretraining Strategy}
  \vspace{-0.2cm}
 \noindent\textbf{A. Contrastive Objective}: The fundamental contrastive objective (shown in Figure \ref{Figure1}(a)), commonly expressed as the InfoNCE loss \cite{oord2018representation}, ensures the alignment between positive sample pairs and augmented views of the same image by maximizing the similarity between positive pairs and minimizing it for negative pairs. The loss (Eq. \ref{eq1}) is calculated as:
 
\begin{equation}
\mathcal{L}_{\text{InfoNCE}} = -\frac{1}{B} \sum_{i=1}^{B} \log \frac{\exp(\mathbf{z}_i \cdot \mathbf{z}_i^+ / \tau)}{\sum_{j=1, j \neq i}^{B} \exp(\mathbf{z}_i \cdot \mathbf{z}_j / \tau)},
\label{eq1}
\end{equation}

Here $\mathbf{z}_i$ represents the embedding of a query sample, while $\mathbf{z}_i^+$ is its positive counterpart. The negative samples, $\{\mathbf{z}_j\}$, are the embeddings of other samples in the batch $B$. The hyperparameter $\tau$ (temperature) controls the separation and smoothness of the learnt representations. 
\vspace{0.7mm}

\noindent\textbf{B. MIM Objective}: This strategy (shown in Figure \ref{Figure1}(b)) entails masking random segments of an input image, requiring the model to reconstruct the missing areas in pixel space. The reconstruction task in MIM is directed by a loss function that assesses the precision of the predicted patches in comparison to their original versions. For a batch of $B$ images, the loss function is defined in Eq. \ref{eq2}:

\begin{equation}
\mathcal{L}_{\text{MIM}} = -\frac{1}{B} \sum_{i=1}^{B} \log f_\theta (\mathbf{x}_i^M | \mathbf{\hat{x}}_i^M),
\label{eq2}
\end{equation}

For each image $i$, $\mathbf{x}_i^M$ refers to the masked patches from the input image, while $\mathbf{\hat{x}}_i^M$ represents the reconstructed versions of those patches. The function $f_\theta$ models the relationship between the visible and masked regions of the input, with the loss being computed as the negative log-likelihood. This formulation ensures that the reconstruction focuses only on the masked areas, encouraging the model to learn meaningful spatial and contextual patterns from the visible regions.

\vspace{0.7mm}
\noindent\textbf{C. MAE \cite{he2022masked} Strategy}: Figure \ref{Figure2} depicts the architecture of MAE. The input image \( X \in \mathbb{R}^{H \times W \times C} \) is divided into a collection of non-overlapping patches \( x \in \mathbb{R}^{N \times (P^2 C)} \). Approximately 75\% of these patches are randomly masked. The remaining visible patches, \( \tilde{x} \), are then processed by a transformer encoder \cite{dosovitskiy2020image} \( f_\theta(\cdot) \), which extracts latent features. The encoded features, combined with placeholders representing the masked patches, are fed into the transformer decoder \( g_\omega(\cdot) \). The decoder’s objective is to reconstruct the pixel values of the original image, ensuring the model accurately captures the spatial and contextual relationships within the input data. The decoder serves just in the pre-training phase, while the encoder is refined for subsequent tasks. The reconstruction is guided by the mean squared error (MSE) loss (Eq. \ref{eq3}), defined as:

\begin{equation}
\mathcal{L}_{\text{MSE}}(x_m, \hat{x}) = \|x_m - \hat{x}\|^2,
\label{eq3}
\end{equation}

where \( x_m \) denotes the original masked patches, and \( \hat{x} \) represents the reconstructed patches. This loss ensures that the reconstructed output \( \hat{x} \) closely approximates the original masked patches, enabling the model to effectively learn spatial and contextual relationships within the image. 

\vspace{-5pt}

\subsection{Downstream RS Tasks}
 \vspace{-0.2cm}
The primary application of MAEs in RS is the self-supervised learning of ViTs on a significant amount of pre-trained RS models \cite{reed2023scale, cong2022satmae, muhtar2023cmid, sun2022ringmo, wang2022advancing, cha2023billion, wang2023ssl4eo, prexl2024senpa}. SSL4EO-S12 \cite{wang2023ssl4eo} demonstrates the use of standard MAE architecture for scene classification tasks. Similarly, Cha et al. \cite{cha2023billion} applied masked image modeling techniques, incorporating parallel connections between multi-head self-attention mechanisms and feed-forward layers of ViT \cite{dosovitskiy2020image}, to address object detection and semantic segmentation challenges. In the framework CMID \cite{muhtar2023cmid}, contrastive learning is integrated with MIM in a self-distillation manner to achieve a balance between global semantic discrimination and localized spatial awareness, for reconstruction tasks. Scale-MAE \cite{reed2023scale} redefines the learning objective of the vanilla MAE by reconstructing high-frequency and low-frequency features of remote sensing imagery, employing a ground sample distance-based positional encoding to improve performance. Wang et al. \cite{wang2022advancing} presented a novel method that uses rotational and variable-size window attention to improve object representation learning. In SatMAE \cite{cong2022satmae}, temporal embeddings are combined with image masking techniques across time, allowing the model to capture spatio-temporal relationships effectively. Table \ref{table1} \cite{wang2024scaling} summarizes the comparison of MAE-based models with contrastive learning-based models across a range of downstream tasks. 
\vspace{-5pt}
\section{Summary and Future Scope}
\vspace{-5pt}
This survey emphasizes the transformative potential of MIM frameworks in addressing the unique challenges of remote sensing, such a spectral diversity, and complex spatial relationships. As MIM evolves, future advancements are expected to focus on integrated architectures, such as joint embedding predictive models, which unify generative reconstruction with discriminative feature alignment. 

\vspace{-5pt}

 \bibliographystyle{IEEEbib}
{footnotesize
}
\end{document}